\documentclass[letterpaper,conference]{IEEEtran}
\IEEEoverridecommandlockouts
\usepackage{graphics} 
\usepackage[linesnumbered,ruled]{algorithm2e}
\graphicspath{{img/}}
\usepackage{epsfig} 
\usepackage{amsmath} 
\usepackage{amssymb}  
\usepackage{amsthm}
\usepackage{url}
\usepackage{bigints}
\usepackage{pifont}
\def\-{\raisebox{.75pt}{-}}
\usepackage{mathtools}
\usepackage{cite}
\usepackage[table]{xcolor}
\usepackage[makeroom]{cancel}
\usepackage{xcolor} 

\usepackage[center]{subfigure}
\usepackage{multirow}
\usepackage{booktabs}
\usepackage{subfigure}
\usepackage{epstopdf}
\usepackage{textcomp}
\usepackage{bm}
\usepackage{caption}
\usepackage{algpseudocode,algorithm} 
\usepackage{color}
\usepackage{mathptmx}
\usepackage[11pt]{moresize}
\usepackage{hhline}
\usepackage{tikz}
\DeclareMathAlphabet{\mathcal}{OMS}{cmsy}{m}{n}
\usepackage{textcomp}
\def\BibTeX{{\rm B\kern-.05em{\sc i\kern-.025em b}\kern-.08em
    T\kern-.1667em\lower.7ex\hbox{E}\kern-.125emX}}

\newcommand\copyrighttext{%
\centering \footnotesize \copyright 2019 IEEE, to appear in the 30th IEEE Intelligent Vehicles Symposium. }
\newcommand\copyrightnotice{%
\begin{tikzpicture}[remember picture,overlay]
\node[anchor=south,yshift=10pt] at (current page.south) {\fbox{\parbox{\dimexpr\textwidth-\fboxsep-\fboxrule\relax}{\copyrighttext}}};
\end{tikzpicture}%
}

\begin{document}
\title{\LARGE Deep Active Learning for Efficient Training of a LiDAR 3D Object Detector
\thanks{$^1$ Robert Bosch GmbH, Corporate Research, Driver Assistance Systems and Automated Driving, 71272 Renningen, Germany.}
\thanks{$^2$ Institute of Measurement, Control and Microtechnology, Ulm University, 89081 Ulm, Germany.}
\thanks{$^3$ School of Electrical Engineering and Computer Science, KTH Royal Institute of Technology, 100 44 Stockholm, Sweden.}
}

\author{Di Feng$^{1,2}$, Xiao Wei$^{1,3}$, Lars Rosenbaum$^1$, 
Atsuto Maki$^3$, Klaus Dietmayer$^2$}

\maketitle

\begin{abstract}
Training a deep object detector for autonomous driving requires a huge amount of labeled data. While recording data via on-board sensors such as camera or LiDAR is relatively easy, annotating data is very tedious and time-consuming, especially when dealing with 3D LiDAR points or radar data. Active learning has the potential to minimize human annotation efforts while maximizing the object detector's performance. In this work, we propose an active learning method to train a LiDAR 3D object detector with the least amount of labeled training data necessary. The detector leverages 2D region proposals generated from the RGB images to reduce the search space of objects and speed up the learning process. Experiments show that our proposed method works under different uncertainty estimations and query functions, and can save up to $60\%$ of the labeling efforts while reaching the same network performance.  
\end{abstract}

\copyrightnotice

\setlength{\parskip}{3mm plus3mm minus3mm}
\setlength{\belowdisplayskip}{8pt} \setlength{\belowdisplayshortskip}{8pt}
\setlength{\abovedisplayskip}{8pt} \setlength{\abovedisplayshortskip}{8pt}

\section{Introduction}\label{sec:introduction}
In recent years deep learning has set the benchmark for object detection tasks on many open datasets (e.g. KITTI~\cite{geiger2012we}, Cityscapes~\cite{Cordts2016Cityscapes}), and has become the de-facto method for perception in autonomous driving. Despite its high performance, training a deep object detector usually requires a huge amount of labeled samples. The annotation process is tedious and time-consuming work, especially for 3D LiDAR points (as discussed in~\cite{lee2018leveraging}), necessitating the development of methods to reduce labeling efforts. Furthermore, a common way to optimize a deep object detector is to feed all training samples into the network with random shuffling. However, the informativeness of each training sample differs, i.e. some are more informative and contribute more to the performance gain, while some others are less informative. A more efficient training strategy is to optimize the network with only the most informative samples. This is specifically helpful when adapting an object detector to new driving scenarios which are different from the previous training set, e.g. from highway to urban scenarios.

Active learning is a training strategy to reduce human annotation efforts while maximizing the performance of a machine learning model (usually in a supervised-learning fashion)~\cite{settles2012active}. In active learning, a model iteratively evaluates the informativeness of unlabeled data, selects the most informative samples to be labeled by human annotators, and updates the training set with the newly-labeled data. Active learning has long been applied to Support Vector Machines (SVM) or Gaussian Processes (GP)~\cite{tong2001support,kapoor2007active,kaboli2017tactile}, and has only recently been used in deep learning for classification of medical images~\cite{gal2017deep} or hyperspectral images in remote sensing~\cite{haut2018active}, 2D image detection~\cite{RoyUN18,kao2018localization}, road-scene image segmentation~\cite{mackowiak2018cereals}, and natural language processing~\cite{siddhant2018deep}.

\begin{figure}[tbp]
	\centering
	\begin{minipage}{1\linewidth}
		\centering
          \includegraphics[width=1\linewidth]{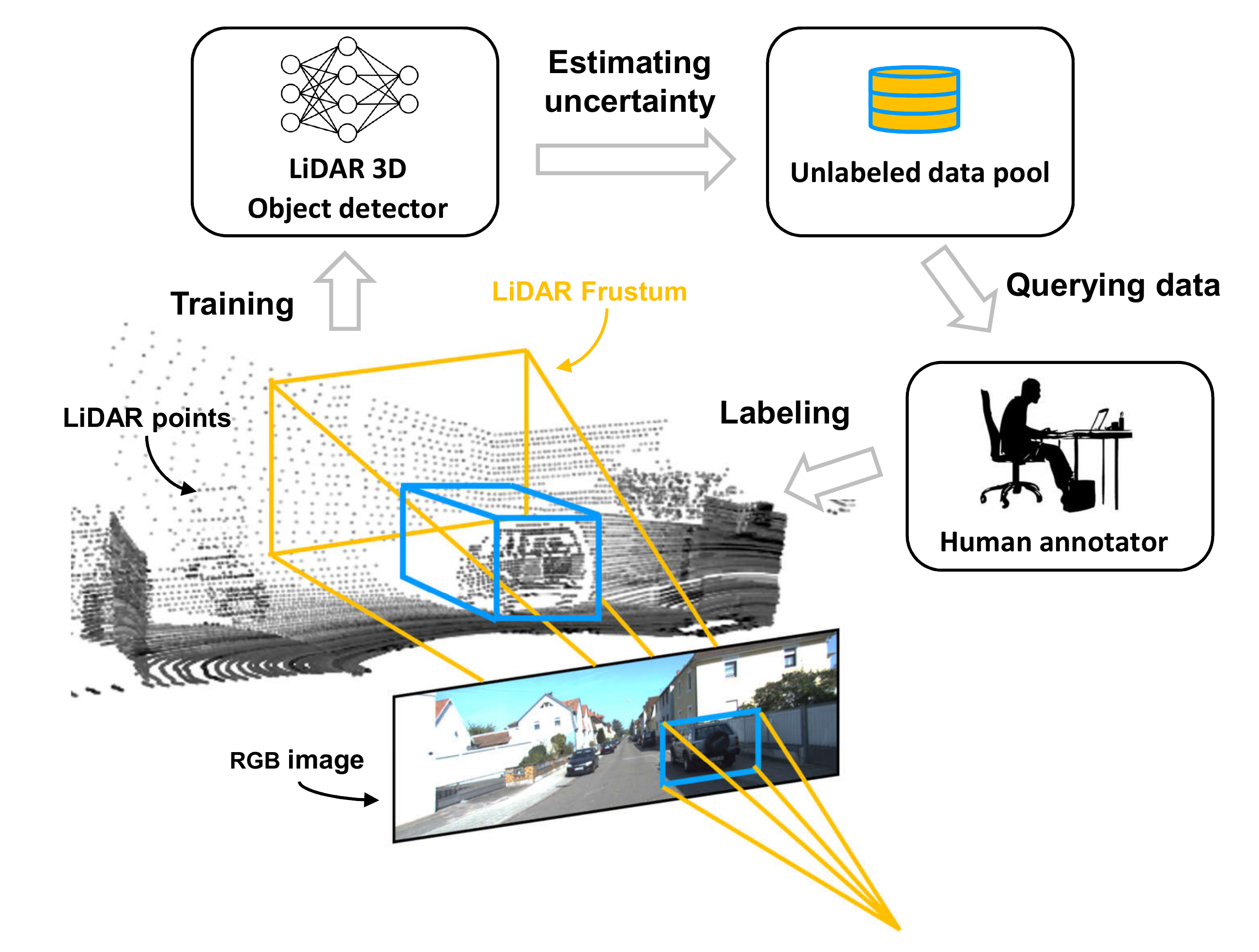}
	\end{minipage}
	\caption{Our proposed active learning method to efficiently train a LiDAR 3D object detector. The detector is based on 2D proposals from images, which serve as seeds to locate objects as frustums in LiDAR 3D space. We assume that there exists a large unlabeled data pool of LiDAR point clouds. The object detector iteratively uses predictive uncertainty to quantify the informativeness  of each sample in the unlabeled data pool, queries the human annotator for the class label and 3D geometrical information of objects, and updates the training set with the newly-labeled data. We validate our method both with ``perfect" image proposals provided by human annotators, or by an on-the-shelf pre-trained image detector with high recall rate.}\label{fig:al_loop}
    \vspace{0.2em}
\end{figure}

In this work, we propose an active learning method to efficiently train a LiDAR 3D object detector for autonomous driving, as in Fig.~\ref{fig:al_loop}. We assume that there exists a large unlabeled data pool of LiDAR point clouds, because it is relatively easy to collect and prepare LiDAR data using a test vehicle. We use the network's predictive uncertainty to quantify the informativeness of each sample in the unlabeled data pool, and assume that the network can iteratively query the human annotator for the class label and 3D geometrical information of objects. Furthermore, as it is much easier to do human labeling with 2D RGB images than 3D point clouds and a lot of pre-trained image detectors with high recall rate exist (e.g. Detectron~\cite{Detectron2018}), we propose to leverage an image detector to provide 2D object proposals which serve as seeds to locate objects, so that the human annotator only needs to label LiDAR points within frustums (see Fig.~\ref{fig:al_loop}). In this way, the 3D labeling efforts can be further reduced and the speed of learning process can be increased. In the experiments, we evaluate our method either by assuming a ``perfect" image detector which provides accurate object proposals, or by an on-the-shelf pre-trained image detector. Results show our method outperforms the baseline method in both experimental settings.      

Our \textbf{contributions} can be summarized as follows: (1) We propose a deep active learning method to significantly reduce the labeling efforts for training a 3D object detector using LiDAR points. To our knowledge, ours is the first attempt to introduce deep active learning for the 3D environment perception on autonomous driving. (2) Our method leverages the 2D object proposals from RGB images, which reduces the search space of objects of interests and speeds up the learning process. (3) We compare several approaches for quantifying uncertainties in the neural network, and study their efficiencies to query informative unlabeled data.

\section{Related Works}\label{sec:related_works}
In this section, we briefly summarize existing works on deep object detection for autonomous driving using LiDAR points as well as deep active learning.

\subsection{Object Detection for Autonomous Driving using LiDAR points}
Most driverless cars are equipped with multiple sensors, such as cameras and LiDARs. Therefore, many methods have been proposed for object detection using camera images~\cite{chen2016monocular,mousavian20173d,teichmann2018multinet}, LiDAR point clouds~\cite{li20163d,li2016vehicle,zhou2017voxelnet,engelcke2017vote3deep,feng2018leveraging}, or the fusion of both to exploit their complementary properties~\cite{chen2016multi,ku2017joint,xu2017pointfusion,qi2017frustum}. 

State-of-the-art deep object detection networks follow two pipelines: the two-stage and the one-stage object detections. In the former pipeline, several object candidates called regions of interest (ROI) or region proposals (RP) are extracted from a scene. Then, these candidates are verified and refined in terms of classification scores and locations. For example, Asvadi \textit{et al.}~\cite{asvadi2017depthcn} cluster LiDAR points for on-ground obstacles using DBSCAN. These clusters are then fed into a ConvNet for 2D detection. Chen \textit{et al.}~\cite{chen2016multi} propose to generate 3D ROIs from the bird’s eye view LiDAR feature maps by a Region Proposal Network (RPN), and combine the regional features from
the front view LiDAR feature maps and RGB camera images
for 3D vehicle detection. In the latter pipeline, single-stage and unified CNN models are used to directly map the input features to the detection outputs. Li \textit{et al.}~\cite{li2016vehicle} and Yang \textit{et al.}~\cite{Yang_2018_CVPR} employ the Fully Convolutional Network (FCN) on LiDAR point clouds to produce an objectness map and several bounding box maps. Caltagirone \textit{et al.}~\cite{caltagirone2017fast} use a FCN for road detection. In this work, we follow the 2-stage object detection pipeline: 2D region proposals are provided by camera images, and the network detects objects using the LiDAR points in the corresponding frustum (Fig.~\ref{fig:al_loop}), similar to \cite{qi2017frustum}. 

\subsection{Deep Active Learning}
Active learning has a long history in the machine learning community (a comprehensive survey is provided by~\cite{settles2012active}), and has been introduced in deep neural networks in $2015$~\cite{gal2017deep}. While many works exist in image classification~\cite{gal2017deep,Beluch_2018_CVPR,haut2018active} and segmentation problems~\cite{mackowiak2018cereals,gorriz17,chitta2018largescale}, little attention has been paid to 2D image detection~\cite{RoyUN18,kao2018localization}. Compared to these works, ours is the first attempt for active learning in 3D object detection problem.

There are numerous approaches to querying unlabeled data, such as variance reduction, query-by-committee, and expected model change~\cite{settles2012active}. Among them, the uncertainty-based approach suggests to use the predictive uncertainty to represent the data informativeness, and to query samples with the highest uncertainty. The effectiveness of this strategy is naturally dependent on obtaining reliable uncertainty estimates. Many recent methods have focused on obtaining such estimates in an efficient manner in deep neural networks. For example, Lakshminarayanan \textit{et al.}~\cite{lakshminarayanan2017simple} propose to use an ensemble of networks to predict uncertainty. Kendall \textit{et al.}~\cite{kendall2017uncertainties} decompose predictive uncertainty into model dependent (epistemic) and data dependent (aleatoric) uncertainties in Bayesian Neural Network. The former is obtained by Monte-Carlo dropout sampling~\cite{Gal2016Uncertainty}, while the latter by predicting the noise in the input data. Guo \textit{et al.}~\cite{GuoPSW17} use a simple calibration technique to improve the network's probability output. Application of these uncertainties has also featured in many recent works. For example, Miller \textit{et al.}~\cite{miller18dropout} employ epistemic uncertainty for object detection in open-set scenarios. Feng \textit{et al.}~\cite{feng2018towards} evaluate the uncertainty estimation in a LiDAR 3D object detection network, and leverage the aleatoric uncertainty to significantly improve the network's robustness against noisy data~\cite{feng2018leveraging}. Ilg \textit{et al.}~\cite{ilg2018uncertainty} compare several uncertainty estimation methods in optical flow. In this work, we use Monte-Carlo dropout~\cite{Gal2016Uncertainty} and Deep Ensembles~\cite{lakshminarayanan2017simple} to estimate uncertainties, and compare their efficiencies in query functions.

\section{Methodology}
\label{sec:method}

In this work, we propose an active learning method to iteratively train a 3D LiDAR detector using the fewest number of labeled training samples, given a large unlabeled data pool. 

Denote the large unlabeled data pool as $\mathcal{D}^u$, which consists of $N^u$ i.i.d data samples $\mathcal{D}^u = \{\mathbf{x}^u_n\}_{n=1}^{N^u}$, and the labeled training dataset as $\mathcal{D}^l$, with $N^l$ samples ($N^u \gg N^l$) and their labels $\mathcal{D}^l = \{\mathbf{x}^l_n, \mathbf{y}^l_n\}_{n=1}^{N^l}$. Also denote the human annotator as $A$ and the object detection model as $M$. Our method is summarized in Alg.~\ref{alg:active_learning}. The keys to the approach are the uncertainty estimation in neural networks, and the data query functions.
 \subsection{Process}
To start the active learning process, the network is initialized with a small labeled dataset $\mathcal{D}^l$ and trained in loop. After each training, the detector evaluates the informativeness of each sample in the unlabeled dataset $\mathcal{D}^u$ by predicting the uncertainty (Step $2$ in Alg.~\ref{alg:active_learning}), selects the most informative samples via a query function (Step $3$), and asks the human annotator for their class labels and object positions (Step $4$). Afterwards, these samples are added to the labeled dataset, and the detector is re-trained. The process iterates until a stop condition is satisfied, e.g. the network's performance converges for several iterations, or a desired performance is reached.

\begin{algorithm}[!t]
	\caption{Active Learning for 3D LiDAR Object Detector}
	\label{alg:active_learning}
	\SetKwInOut{Input}{Input}
	\SetKwInOut{Output}{Output}	
	
	\Input{$\mathcal{D}^u, \mathcal{D}^l, A$} \;
	\textbf{Initialization:} $M \gets trainDetector(\mathcal{D}^l)$ \\
	\While{$notStopCondition()$} 
	{
		$\mathcal{U}(\mathcal{D}^u) \gets uncertaintyEstimate(\mathcal{D}^u) $ \;

		$ X^{\ast}_u \gets dataQuery \big(\mathcal{D}^u,\mathcal{U}(\mathcal{D}^u) \big)$ \Comment{\small A subset of unlabeled data}\;  

		$Y^{\ast} \gets dataLabel(X^{\ast}_u, A)$ \Comment{\small Class label and object location}\;

$\mathcal{D}^l \gets \mathcal{D}^l \cup \{X^{\ast}_u,Y^{\ast}\}$ \Comment{\small Add data to the training dataset}\;

$\mathcal{D}^u \gets \mathcal{D}^u\backslash X^{\ast}_u$ \Comment{\small Delete data from the unlabeled dataset}\;

$M \gets trainDetector(\mathcal{D}^l)$ \Comment{\small Update the network}
	}
\end{algorithm}
\subsection{3D Object Detector}
\begin{figure}[tbp]
	\centering
	\begin{minipage}{1\linewidth}
		\centering
          \includegraphics[width=0.96\linewidth]{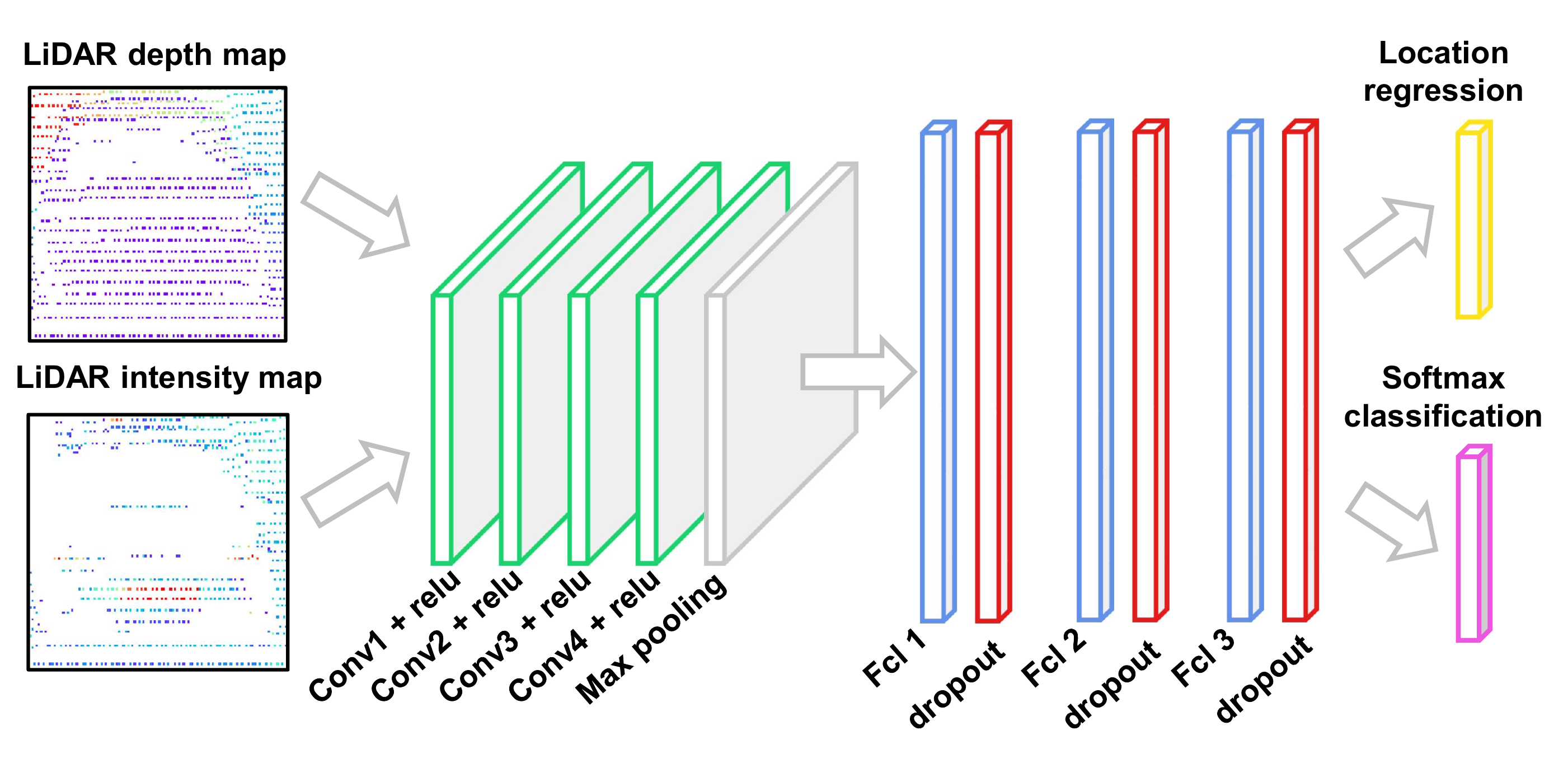}
	\end{minipage}
	\caption{Network architecture. The detector takes the LiDAR depth and intensity maps as input and outputs the objectness score and object location information (width, length, height, and depth).}\label{fig:network}
    \vspace{-1.2em}
\end{figure}
\subsubsection{Inputs and Outputs}
As mentioned in the Introduction (Sec.~\ref{sec:introduction}), our detector leverages the 2D region proposals in RGB images, which build frustums in the 3D space. We project those LiDAR points in frustums onto the front-view camera plane and build sparse LiDAR depth and intensity maps. Since these maps bring complementary LiDAR information (see Fig.~\ref{fig:network}), we concatenate them to build the network inputs. Note that we do not perform interpolation (e.g. Delaunay Triangulation \cite{asvadi2017depthcn} and Bilateral Filtering \cite{premebida2016high}) in order to avoid interpolation artifacts. The network outputs softmax classification scores $\mathbf{s}$, and object locations $\mathbf{t}$. We encode an object's 3D position as the relative width $\hat{w}=\frac{w}{w_{max}}$, length $\hat{l}=\frac{l}{l_{max}}$, and height $\hat{h}=\frac{h}{h_{max}}$ of the bounding box, as well as the euclidean distance between our ego-vehicle and the object centroid $\hat{d}=\frac{d}{d_{max}}$, i.e. $\mathbf{t} = \{\hat{w},\hat{l},\hat{h},\hat{d}\}$. We select $w_{max}, l_{max}, h_{max}, d_{max}$ based on the heuristics from the dataset.  

\subsubsection{Network Architecture}
Our object detector is built on the ConvNet depicted in Fig.~\ref{fig:network}. It is composed of four convolutional layers (each with $32$ $3\times3$ kernels and relu activation), a pooling layer, three fully connected layers (each with $256$ hidden units), and three dropout layers. The dropout layers are used for stochastic regularization during training and uncertainty estimation during testing.

\subsection{Uncertainty Estimation and Query Functions}
\subsubsection{Uncertainty Estimation}
In this work, we use the predictive probability $p(\mathbf{y}|\mathbf{x})$ in classification to estimate uncertainty in our object detection network. For simplification, we denote $\mathbf{x}$ as a data point and $\mathbf{y}$ classification labels. A direct way to obtain predictive probability is by softmax output, i.e. $p(\mathbf{y}|\mathbf{x})=softmax(\mathbf{x})$. However, as discussed in several works (e.g.~\cite{Gal2016Uncertainty},\cite{Beluch_2018_CVPR}), the softmax output may assign high probability to unseen data, resulting in over-confident predictions. Therefore, we also use two recent methods to obtain uncertainty estimates, namely, Monte-Carlo dropout (MC-dropout \cite{Gal2016Uncertainty}) and Deep Ensembles (\cite{lakshminarayanan2017simple}).

MC-dropout~\cite{Gal2016Uncertainty} regards dropout regularization as approximate variational inference in the Bayesian Neural Network framework, and extracts predictive uncertainty by performing multiple feed-forward passes with dropout active during test time. More specifically, given a test point $\mathbf{x}$, the network performs $T$ inferences with the same dropout rate as training, and averages the outputs to approximate the predictive probability:
\begin{equation}\label{equ:prob1}
p(\mathbf{y}|\mathbf{x})\approx \frac{1}{T}\sum_{t=1}^T p(\mathbf{y}|\mathbf{x},\mathbf{W}_t) = \frac{1}{T}\sum_{t=1}^T softmax_{(\mathbf{W}_t)}(\mathbf{x}),
\end{equation}
with $\mathbf{W}_t$ being network's weights for the $t^{th}$ inference. 

Compared to MC-dropout, Deep Ensembles~\cite{lakshminarayanan2017simple} estimates predictive uncertainty in an non-Bayesian way. It suggests to train several networks with the same architecture but with random initialization, and average the networks' outputs during testing. Let $E$ be the number of ensembles and $M_e$ a single network in the ensemble, similar to Eq.~\ref{equ:prob1}, we have:
\begin{equation}\label{equ:prob2}
p(\mathbf{y}|\mathbf{x}) \approx \frac{1}{E}\sum_{e=1}^E p(\mathbf{y}|\mathbf{x},\mathbf{M}_e)  = \frac{1}{E}\sum_{e=1}^E softmax_{(\mathbf{M}_e)}(\mathbf{x}).
\end{equation}

\subsubsection{Query Functions}
Based on the above-mentioned methods to obtain the predictive probability, we can calculate the informativeness (or uncertainty) for each sample in the unlabeled data pool $\mathcal{D}^u$ and use acquisition functions to query the most uncertain samples. A common way is to measure the Shannon Entropy (SE)~\cite{shannon2001mathematical} with:
\begin{equation}
\mathcal{H}[\mathbf{y}|\mathbf{x}] = -\sum_{c=1}^{C} p(y=c|\mathbf{x}) \log p(y=c|\mathbf{x}), 
\end{equation}
and query unlabeled samples with the highest Entropy values. $y$ refers to a classification label, and $C$ the number of classes.

Additionally, since both MC-dropout and Deep Ensembles provide samples from the predictive probability distribution (i.e. $p(\mathbf{y}|\mathbf{x},\mathbf{W}_t)$ or $p(\mathbf{y}|\mathbf{x},\mathbf{M}_e)$), we can use them to measure the Mutual Information (MI)~\cite{gal2017deep} between the model weights and the class labels, and query unlabeled data with the highest MI. The mutual information using MC-dropout is calculated by:
\begin{equation}
\begin{split}
& \mathcal{I}[\mathbf{y};\mathbf{W}] = \mathcal{H}[\mathbf{y}|\mathbf{x}] - \mathbb{E}_{p(\mathbf{W}|\mathcal{D}^l)} \mathcal{H}[\mathbf{y}|\mathbf{x}, \mathbf{W}] \\
& \approx \mathcal{H}[\mathbf{y}|\mathbf{x}] + \frac{1}{T}\sum_{t=1}^T \sum_{c=1}^{C} p(y=c|\mathbf{x},\mathbf{W}_t) \log p(y=c|\mathbf{x},\mathbf{W}_t),
\end{split}
\end{equation}
where $p(\mathbf{W}|\mathcal{D}^l)$ indicates the posterior distribution of network weights $\mathbf{W}$ given the training dataset $cal{D}^l$. For Deep Ensembles, we only need to replace $W$ with $M$. As discussed in our previous work~\cite{feng2018towards}, SE and MI capture different aspects of uncertainty: SE measures the output uncertainty (predictive uncertainty), whereas MI measures the model's confidence in the data (epistemic uncertainty).
\section{Experimental Results}\label{sec:result}
\subsection{Experimental Design}
We evaluate our proposed method based on two experimental settings. In the first experiment, we study the active learning performance with different uncertainty estimation approaches and query functions. To avoid the influence from the RGB image detector, we assume a ``perfect" image detector that provides only accurate object proposals. This is achieved by extracting objects using their ground-truth labels. Thus, we simplify the object detection problem in this setup to a classification and a location regression problem. In the second experiment, we use a pre-trained image detector to predict region proposals, which contain either object or background images.

Both experiments are conducted on the KITTI dataset~\cite{geiger2012we}. The LiDAR depth and intensity maps are generated by projecting LiDAR points onto the image plane and then warped to $100\times100$ pixels, with additional $5$ pixels padding to include context information, similar to~\cite{girshick2014rich}. To start the active learning process, the network is trained with some samples randomly selected from the training data and balanced over all classes ($200$ samples per class). The network is trained with Adam optimizer, together with dropout and $l_2$ regularization to prevent over-fitting. We set dropout rate to be $0.5$, and the weight decay to be $10^{-4}$. At each query step, $200$ samples are selected from the remaining ``unlabeled'' training dataset, and the network is re-trained from scratch. At each query step, we calculate both the classification accuracy and the mean squared error (MSE) on the test data set, to track classification and localization performance, respectively. To have a fair comparison among different uncertainty estimation approaches, the MC-dropout methods are evaluated by performing the single forward pass on the test dataset without dropout. We also use the predictions from only one network in the ensemble for evaluation. For MC-Dropout, $20$ forward passes are used during inference, while the ensemble consists of $5$ classifiers. We fix the number of query steps to be $60$, and repeat each experiment $3$ times. Besides, we report the performance of a full-trained network.

\subsection{Evaluation with a ``Perfect" RGB Image Detector}
\subsubsection{Setting}
We divide objects into five classes, namely, ``Small Vehicle'' (including ``Car'' and ``Van'' categories in KITTI), ``Human'' (including ``Pedestrian'', ``Person sitting'', and ``Cyclist''), ``Truck'', ``Tram'', and ``Misc''.  We randomly divide the dataset into a training set with $31500$ samples, a test set with $6000$ samples and a val set with $3000$ samples.

We compare the active learning strategy with the baseline method which randomly queries the unlabeled data points following the uniform distribution. We also study the behavior of active learning using five different query functions. ``Softmax + Entropy", ``MC-dropout + Entropy" and ``Ensembles + Entropy" use the same query function that maximizes Shannon Entropy. ``MC-dropout + MI" and ``Ensembles + MI" query the data by maximizing the mutual information.

\subsubsection{Active Learning Performance}
\begin{figure*}[tbp]
	\centering
	\begin{minipage}{1\linewidth}
		\centering
          \includegraphics[width=0.96\linewidth]{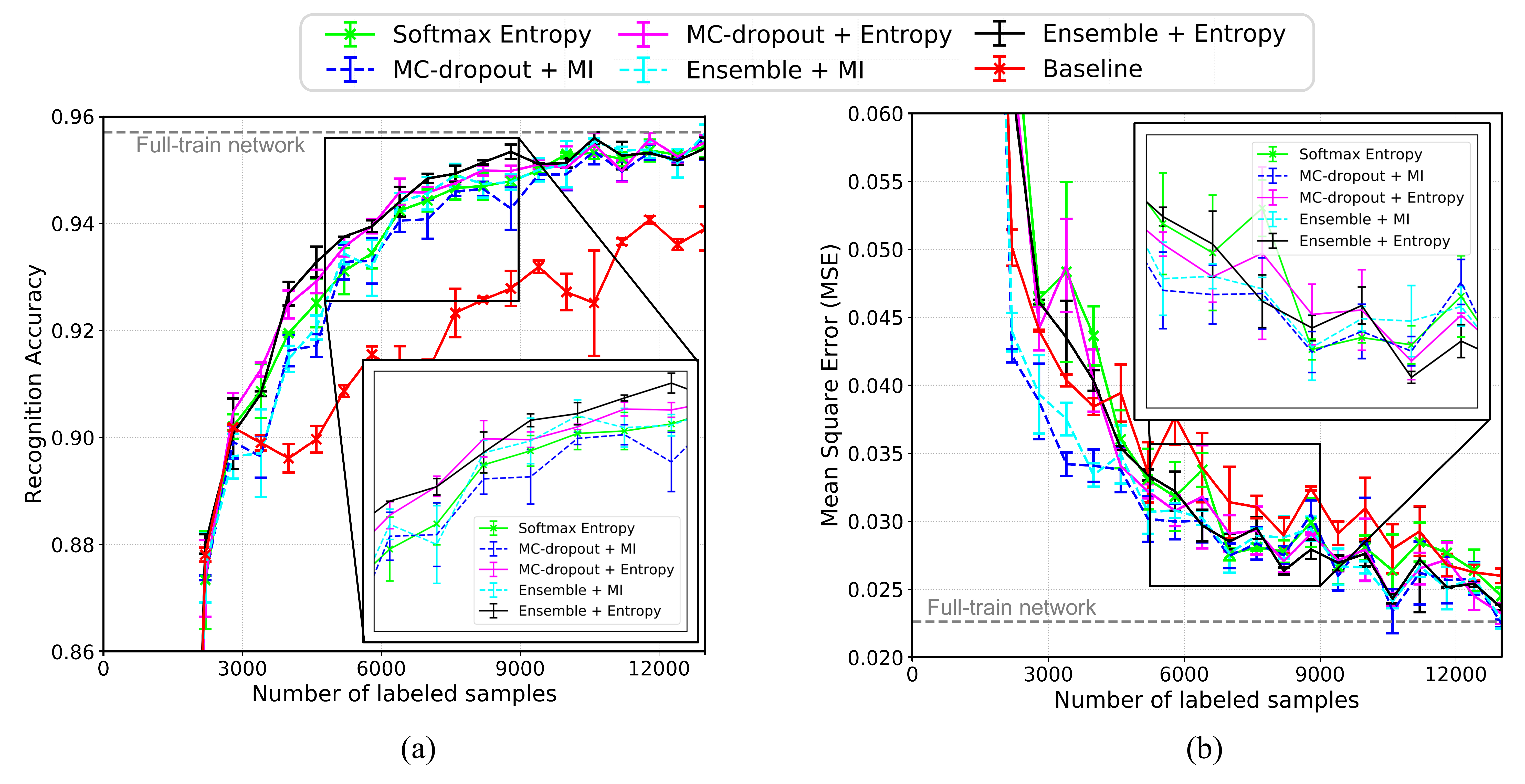}
	\end{minipage}
	\caption{Detection performance for the baseline and several active learning methods with different uncertainty estimations and query functions. The horizontal axis represents the increasing number of labeled training data samples. All networks are initialized with $1000$ samples balanced over all classes. At each query step, $200$ samples are queried from the unlabeled data pool. The vertical axis represents the detection performance for \textbf{(a)} classification and \textbf{(b)} object localization.}\label{fig:al_performance}
    \vspace{-1.0em}
\end{figure*}

Results are shown in Fig.~\ref{fig:al_performance}. We have three observations: (1) All active learning methods significantly outperform the baseline method for classification and localization tasks. They achieve higher recognition accuracy or lower mean squared error with the same number of training data as the baseline method. (2) With a relatively small amount of data (e.g. $<7000$ samples), MI-based query functions (``MC-dropout + MI" and ``Ensembles + MI") consistently perform better than their Entropy counterparts (``MC-dropout + Entropy" and ``Ensembles + Entropy") in the localization task (Fig.~\ref{fig:al_performance}(b)), whereas Entropy-based methods are more suitable for the classification task (Fig.~\ref{fig:al_performance}(a)). (3) Using MC-dropout and Deep Ensembles to estimate uncertainty results in slightly better active learning results compared to using a single softmax output (see ``Softmax + Entropy", ``MC-dropout + Entropy" and ``Ensembles + Entropy").

Tab.~\ref{tab:improvement} compares the number of labeled training samples required to train the detector so that it can reach a certain error level relative to the full-trained network. Denote $accu_{full}$ and $accu_m$ as the classification accuracy from the full-trained detector and the classification accuracy from a detector in the active learning process, respectively. Also denote $mse_{full}$ and $mse_m$ as the mean squared error for localization. We define the relative error for classification as $|accu_{full}-accu_m|$ and for localization as $|mse_m-mse_{full}|/mse_{full}$. We also calculate the percentage of labeling efforts saved by the active learning methods compared to the Baseline. As can be seen in the Table, our methods reach the relative error with significantly fewer training points, saving up to $60\%$ training samples. In addition, using MC-dropout or Deep Ensembles to evaluate predictive uncertainty produces similar or better results than the single softmax approach, indicating that they can better represent the informativeness of unlabeled data.

\begin{figure*}[tbp]
	\centering
	\begin{minipage}{1\linewidth}
    \includegraphics[width=0.96\linewidth]{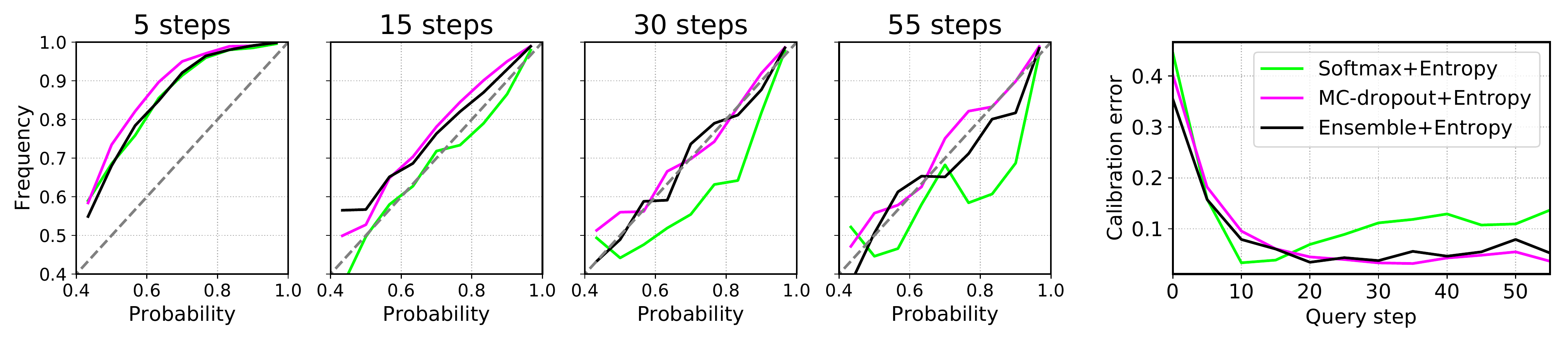}
	\end{minipage} 
	\caption{A comparison of the calibration quality of predictive uncertainty from ``Softmax + Entropy'', ``MC-dropout + Entropy'', and ``Ensemble + Entropy'' averaged over three runs. To this end, we divide the probability values into several bins and calculate the frequency of correct predictions in each bin. The calibration plots at query step $5$, $15$, $30$, and $55$ are shown in the first four subplots. The diagonal line indicates perfect calibration, where the predictive uncertainty matches the observed frequency of correct predictions. The evolution of calibration error w.r.t. query step is shown in the right plot. A smaller error value indicates better uncertainty estimation.}\label{fig:calibration_plot}
    \vspace{-1.0em}
\end{figure*}

\begin{figure*}[tbp]
	\centering
	\begin{minipage}{1\linewidth}
		\centering
          \includegraphics[width=0.96\linewidth]{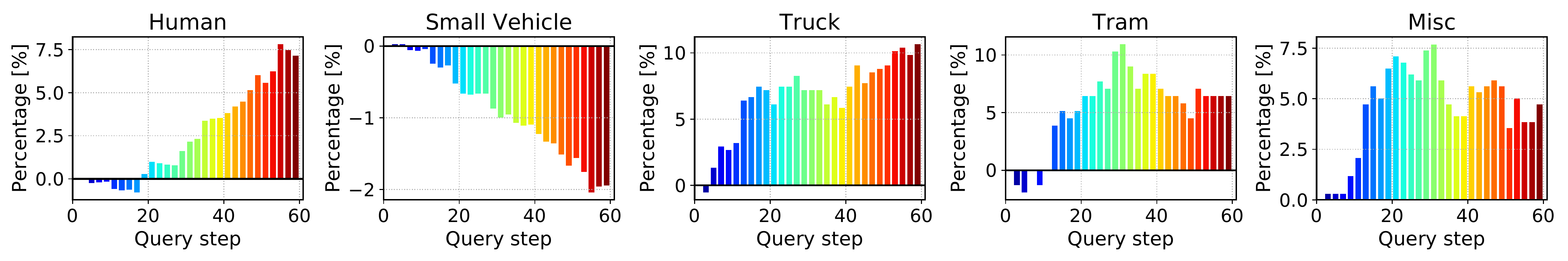}
	\end{minipage}
	\caption{Class distribution of queried samples for ``Ensemble + Entropy" compared to the baseline method. The vertical axis represents the accumulative queried samples relative to the baseline, normalized by the number of samples of the corresponding classes in the unlabeled data pool. Compared to the baseline method, active learning queries fewer samples from ``Small Vehicle'' and more from the other classes. Note that since there are much more objects in ``Small vehicle" than the other classes, a small percentage drop in ``Small vehicle" means a much more percentage rise in other classes. }\label{fig:class_distribution}
    \vspace{-1.0em}
\end{figure*}

\begin{figure}[tpb]
\centering
    \begin{minipage}{0.98\linewidth}
	\subfigure[]{\label{fig:sparsification_plot}\includegraphics[width=0.49\textwidth]{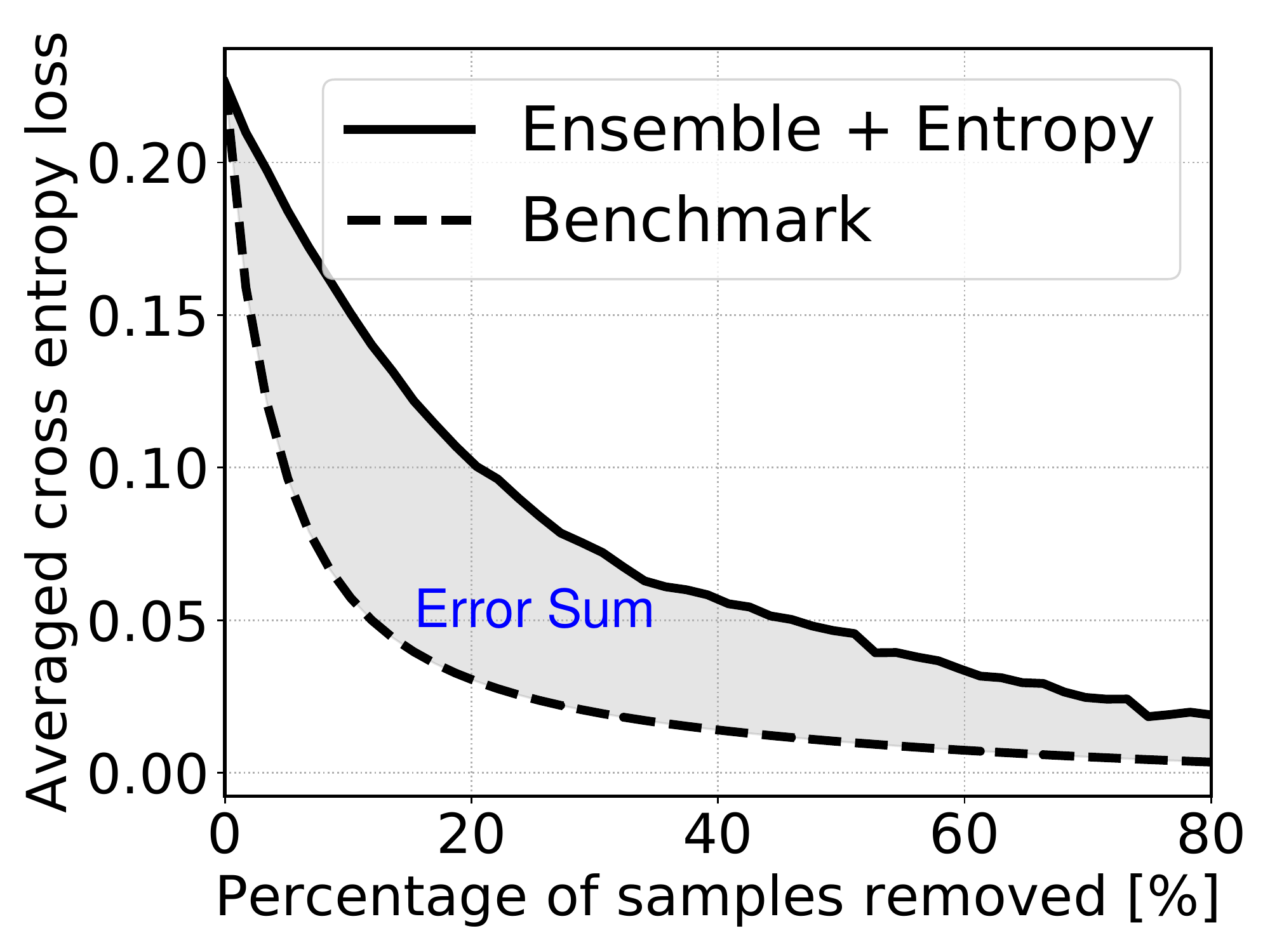}}
	\subfigure[]{\label{fig:sparsification_error}\includegraphics[width=0.49\textwidth]{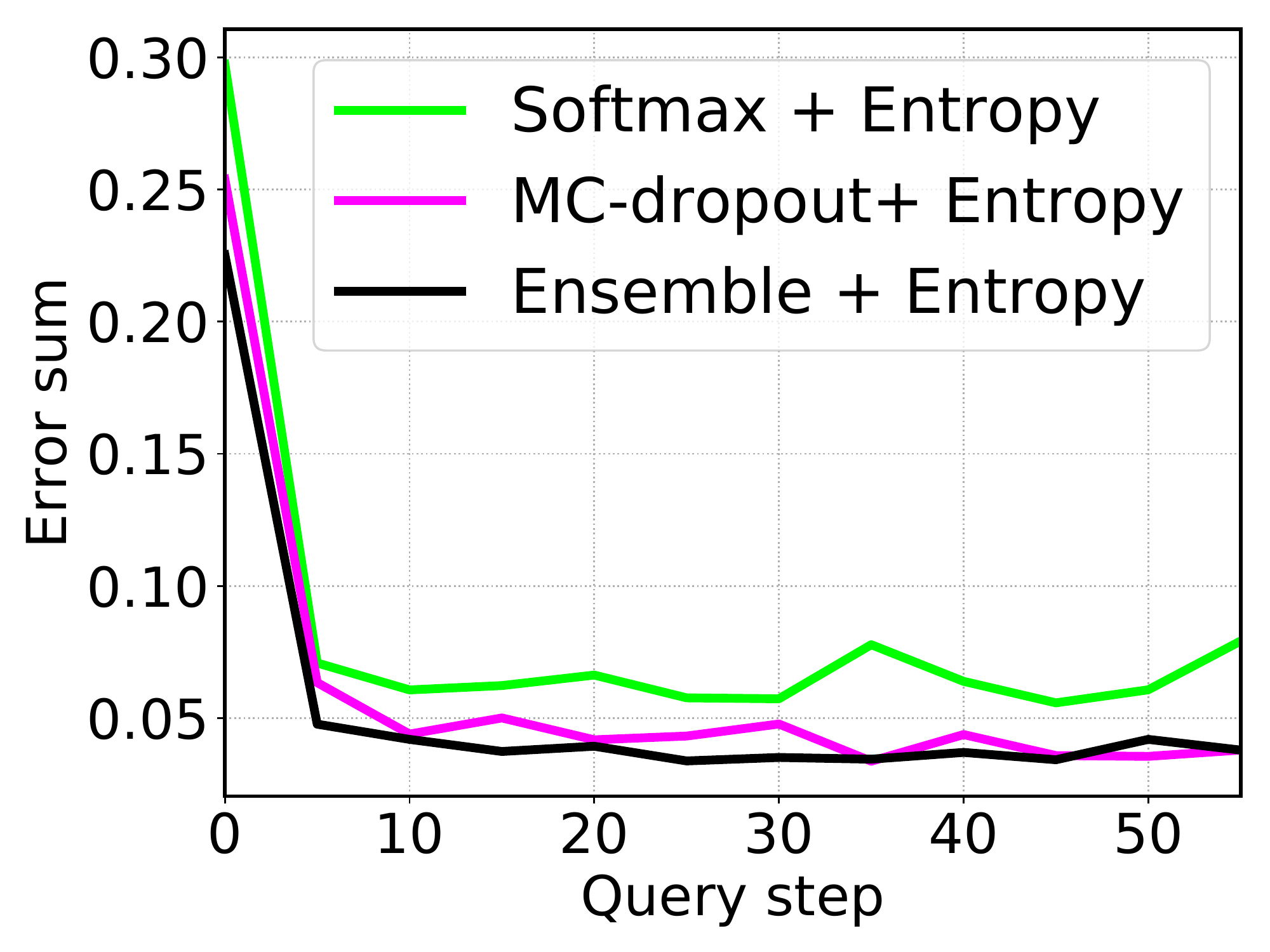}}
	\caption{(\textbf{a}): Error curve for ``Ensemble + Entropy'' at a specific query step. We use the cross entropy loss to represent the error. (\textbf{b}): The evolution of error sum w.r.t. query step. A smaller error sum indicates better uncertainty estimation. The plots are averaged over three runs.}
	\vspace{-1.0em}
\end{minipage}
\end{figure}

 \begin{table*}[tbp]
 \rowcolors{1}{gray!15}{white}
\centering
    \resizebox{0.96\linewidth}{!}{\begin{tabular}{c | c c c c | c c c c}
   & \multicolumn{4}{c|}{Recognition accuracy (Classification)} & \multicolumn{4}{c}{Mean Squared Error (Localization)} \\ 
    Relative error to the full-trained network & $5\%$ & $4\%$ & $3\%$ & $2\%$ & $75\%$ & $30\%$ & $15\%$ & $5\%$ \\  \hline
    Baseline & $5000$ & $6800$ & $9000$ & $11400$ & $4600$ & $6800$ & $10000$ & $12200$   \\ 
	Softmax + Entropy 		& $3200 (\text{+}36\%)$ & $3800 (\text{+}44\%)$ & $4400 (\text{+}51\%)$ & $6000 (\text{+}47\%)$ & $3800 (\text{+}17\%)$ & $6600 (\text{+}3\%)$ & $8800 (\text{+}12\%)$ & $10600 (\text{+}13\%)$ \\
	MC-dropout + MI 		& $3600 (\text{+}28\%)$ & $3800 (\text{+}44\%)$ & $4600 (\text{+}49\%)$ & $6000 (\text{+}47\%)$ & $2000 \mathbf{(\text{+}57\%)}$ & $5400 (\text{+}21\%)$ & $8600 (\text{+}14\%)$ & $10000 \mathbf{(\text{+}18\%)}$ \\
    MC-dropout + Entropy 	& $2600 \mathbf{(\text{+}48\%)}$ & $3600 (\text{+}47\%)$ & $4400 (\text{+}51\%)$ & $5600 (\text{+}51\%)$ & $3800 (\text{+}17\%)$ & $6200 (\text{+}9\%)$ & $8000 (\text{+}20\%)$ & $10200 (\text{+}16\%)$ \\

    Ensemble + MI 			& $3400 (\text{+}32\%)$ & $3800 (\text{+}44\%)$ & $4400 (\text{+}51\%)$ & $6000 (\text{+}47\%)$ & $ 2200 (\text{+}52\%)$ & $4400 \mathbf{(\text{+}35\%)}$ & $9000 (\text{+}10\%)$ & $10200 (\text{+}16\%)$  \\
    Ensemble + Entropy 	& $3000 (\text{+}40\%)$ & $3400 \mathbf{(\text{+}50\%)}$ & $3800 \mathbf{(\text{+}58\%)}$ & $4400 \mathbf{(\text{+}61\%)}$ & $3200 (\text{+}30\%)$ & $4600 (\text{+}32\%)$ & $7600 \mathbf{(\text{+}24\%)}$ & $10600 (\text{+}13\%)$ \\ 
    \end{tabular}}
    \caption{The number of labeled training samples required to train the detector in order to reach a certain relative error to the full-trained network. The table also shows the percentage of labeling efforts saved by the active learning methods compared to the Baseline.}\label{tab:improvement}
 \end{table*}

\subsubsection{Understanding How Active Learning Works}
Our proposed method is based on a good uncertainty estimation. A better uncertainty score can better represent the data informativeness, leading to a better active learning performance. In this regard, we use the calibration plot and error curve to evaluate the quality of predictive uncertainty, similar to~\cite{Beluch_2018_CVPR} and ~\cite{ilg2018uncertainty}. Furthermore, we investigate the class distribution of the queried samples.

\noindent \textbf{Calibration Plot:} 
A calibration plot is a quantile-quantile plot (QQ-plot) between the predictive probability of a model and the observed frequency of correct predictions. A well calibrated uncertainty estimation should match the frequency of correct prediction, showing as a diagonal line. As an example, $60\%$ of samples should be correctly classified as class $c$, when a well-calibrated network predicts them with probability output $p(y=c|\mathbf{x})=60\%$. We compare the uncertainty estimations using single softmax (``Softmax + Entropy"), MC-dropout (``MC-dropout + Entropy"), and Deep Ensembles (``Ensemble + Entropy") on the test dataset. The calibration plots at query steps $5$, $15$, $30$, $55$ are illustrated in the left four figures in Fig.~\ref{fig:calibration_plot}, and the evolution of calibration error with query step in the right figure. The calibration error is calculated as the mean absolute deviation between the frequency and the diagonal line. The figures show that at the first few query steps all methods are ``under-confident" with predictions, as their calibration plots are above the diagonal line. This indicates that networks are under-fitted with only a small number of training samples. As the query step increases, the predictive uncertainties from MC-dropout and Deep Ensembles become well-calibrated. However, the network using single softmax turns out to be over-fitted with data and produces over-confident predictions, resulting in a calibration plot under the diagonal line. The experiment show that MC-dropout and Deep Ensembles produce more reliable uncertainty estimation and improve the learning performance more than the single softmax.   

\noindent \textbf{Error Curve:}
Another way to evaluate the quality of predictive uncertainty is via the error curve (or ``sparsification plot" proposed in~\cite{ilg2018uncertainty}). It is assumed that a well-estimated predictive uncertainty should correlate with the true error, and by gradually removing the predictions with high uncertainty, the average errors over the rest of the predictions will decrease. In our problem, we use the cross entropy loss to represent error. An exemplary error curve for ``Ensemble + Entropy" at a specific query step is shown by Fig.~\ref{fig:sparsification_plot}. The benchmark is obtained by thresholding the predictions by their cross entropy losses (true errors). Note that for each uncertainty estimation method, we obtain a different benchmark. Threrefore, we calculate the mean absolute deviation between the error curve and its benchmark, denoted as ``Error sum", to have a fair comparison on the quality of uncertainty estimation. Fig.~\ref{fig:sparsification_error} illustrates the evolution of the ``Error sum" over the query steps for single softmax, MC-dropout and Deep Ensembles. Both MC-dropout and Deep Ensembles consistently outperform the single softmax method with smaller error sums.

\noindent \textbf{Distribution of Sampled Objects}
To further understand how active learning outperforms the baseline method, we compare the class distribution of queried samples between ``Ensemble + Entropy" and Baseline.  This is calculated by taking the difference between the two at each step, normalized by the total number of samples from the corresponding classes in the unlabeled data pool. The results are shown in Fig.~\ref{fig:class_distribution}. The unlabeled data is highly class-imbalanced with ratios 
``Small Vehicle"$=78\%$, ``Human"$=15.6\%$, ``Truck"$=2.7\%$, ``Tram"$=1.3\%$, ``Misc"$=2.4\%$. However, our method naturally alleviates such problem by querying fewer samples from ``Small Vehicle" and more from other classes. Note that this effect of balancing samples over classes is due to using the uncertainty estimation in the query function rather than an ad-hoc solution that explicitly over/undersample examples.

\subsection{Evaluation with a Pre-trained RGB Image Detector}
\subsubsection{Setting} 
In this experiment, we evaluate the active learning method based on region proposals provided by a RGB image detector. To this end, we follow~\cite{chen2016multi} to divide the KITTI dataset into a \textit{train set} and a \textit{val set}, and use the RGB image detector proposed by~\cite{qi2017frustum}. The \textit{train set} is used to fine-tune the image detector which has been trained on COCO dataset~\cite{lin2014microsoft}, and the \textit{val set} is used to evaluate our method. We consider to detect objects with the classes ``Small vehicle" and ``Human" (same to the previous experiment). A proposal is assigned positive when its 2D Intersection over Union (IoU) with the ground truth is larger than $0.5$. Proposals with IoU smaller than $0.5$ or from other object classes are marked as ``Background". The recall scores of the image detector is shown by Tab.~\ref{tab:recall}.

\begin{table}[h!]
\rowcolors{1}{gray!15}{white}
  \begin{center}
    \begin{tabular}{c | c c} 
       & Small Vehicle & Human \\ \hline
      Recall & $0.917$ & $0.862$ \\
    \end{tabular}
    \caption{Recall rate for the RGB image detector.} \label{tab:recall}
  \end{center}
\end{table}

Based on image proposals, we build a training data pool with $17221$ samples to train our active learning method. These samples are selected with their IoU being either larger than $0.5$ (Positive) or below $0.2$ (Background). The test dataset contains $6000$ samples, including those with IoU ranging between $0.2$ and $0.5$. Note that ignoring samples with some IoU ranges is a common procedure when training object detectors, as discussed in~\cite{chen2016multi,girshick2014rich}.

\subsubsection{Active Learning Performance}
\begin{figure}[tpb]
\centering
    \begin{minipage}{0.98\linewidth}
	\subfigure[]{\label{fig:accuracy_detection}\includegraphics[width=0.49\textwidth]{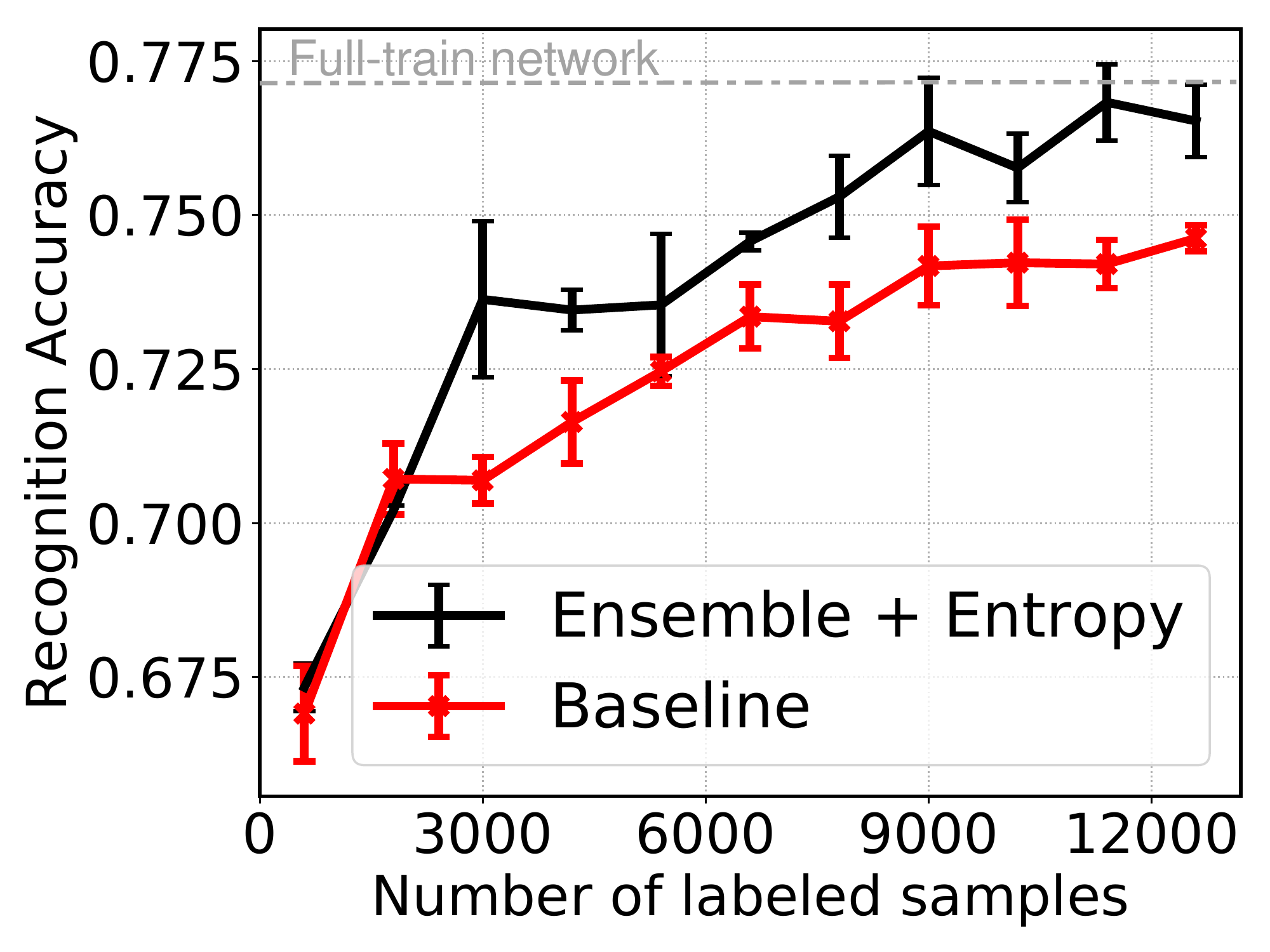}}
	\subfigure[]{\label{fig:mse_detection}\includegraphics[width=0.49\textwidth]{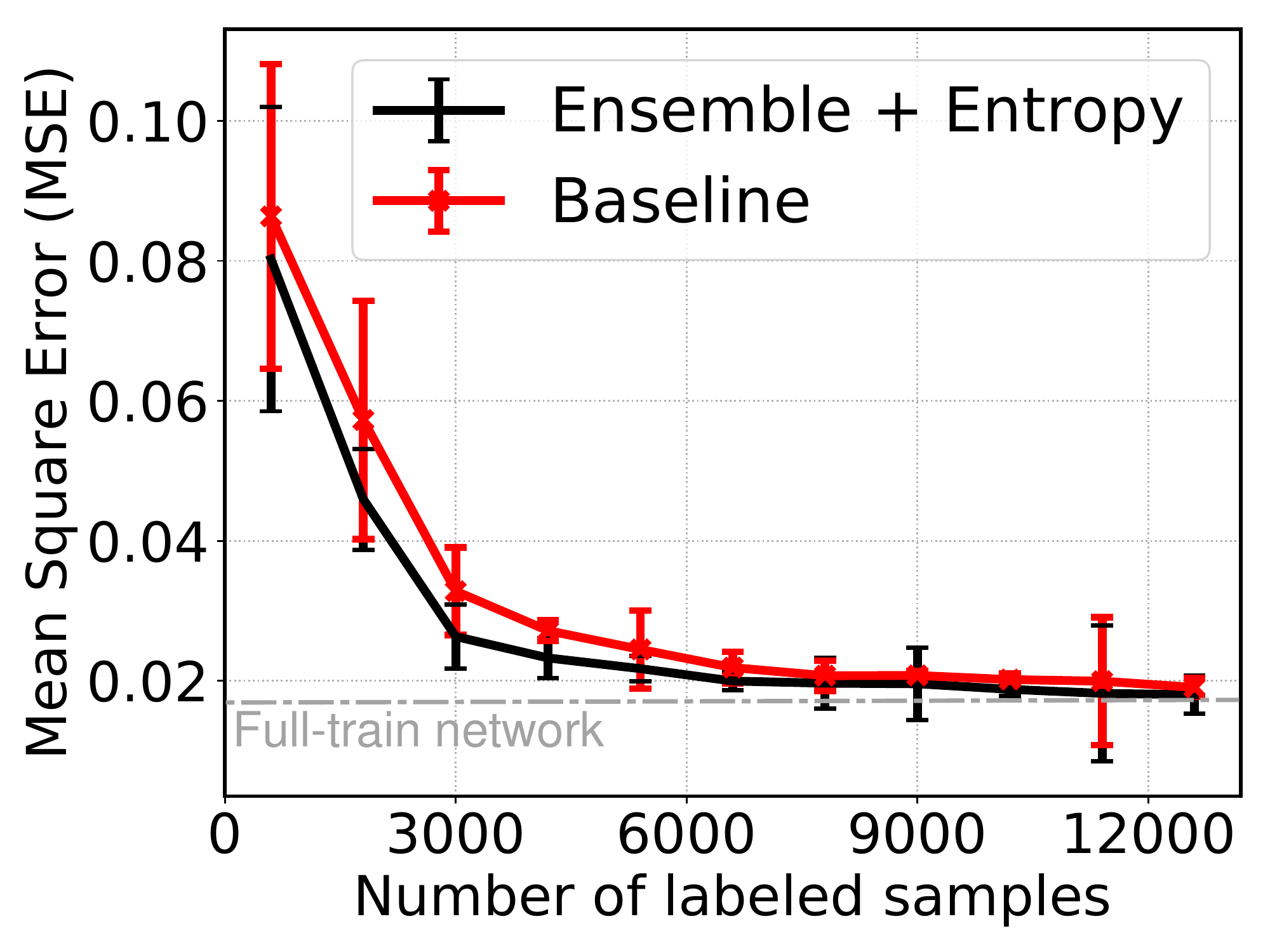}}
	\caption{A comparison of learning learning between active learning and baseline methods for (\textbf{a}) classification and (\textbf{b}) localization tasks. The active learning is built by ``Ensemble + Entropy" strategy.}
	\vspace{-1.0em}
\end{minipage}
\end{figure}

We compare the active learning based on ``Ensemble + Entropy" strategy with the baseline method. Results are shown in Fig.~\ref{fig:accuracy_detection} and Fig.~\ref{fig:mse_detection}. Compared to the first experiment (Fig.~\ref{fig:al_performance}), the network in this experiment results in a lower recognition accuracy, as this experiment is more challenging than the previous one. Despite of this, active learning consistently outperforms the baseline methods by reaching the same recognition accuracy or mean squared error with fewer labeled training samples. 

\section{Discussion and Conclusion}
\label{sec:conclusion}
We presented a method that leverages active learning to efficiently train a LiDAR 3D object detector. The network predicts the object classification score and 3D geometric information based on 2D proposals on camera images. We conducted experiments using a ``perfect" image detector to compare several ways of uncertainty estimation and query functions. Results show that MC-dropout and Deep Ensembles provide more reliable predictive uncertainties compared to the single softmax output, and achieve better active learning performance. We also used a pre-trained image detector to predict image region proposals. In both experimental settings, our active learning method reaches the same detection performance with significantly fewer training samples compared to the baseline method, saving up to $60\%$ labeling efforts.

We show that building query functions based on predictive uncertainty in classification is effective not only in improving recognition accuracy, but also in reducing the mean squared error for the localization task at the same time (e.g. Tab.~\ref{tab:improvement} and Fig.~\ref{fig:al_performance}). This indicates that by sharing weights in the hidden layers, the classification and localization are related to each other in the object detection network. It is an interesting future work to introduce location uncertainty into our active learning method. Furthermore, in this work the network is retrained from scratch after each query step. In applications where we want to adapt an object detector to new driving scenarios (as discussed in Sec.~\ref{sec:introduction}), it is preferable to fine-tune the network with newly-labeled data. Employing our proposed active learning method in such a ``life-long learning'' scenario is an interesting future work. 

One limitation of our method is that the performance of the LiDAR detector is highly dependent on region proposals from the image detection. Despite on-the-shelf image detectors already achieve high detection performance, the LiDAR detector can not handle false negatives in images. In order to guarantee highly qualified region proposals, we can incorporate the image detector into the active learning loop, i.e. the region proposals are first provided by the image detector and then corrected by human annotators. We leave this as an interesting future work.

\section*{Acknowledgment}
We thank William H. Beluch, Radek Mackowiak, and Christian H. Schuetz for the suggestions and fruitful discussions.  

\bibliographystyle{IEEEtran}
\bibliography{bibliography}

\end{document}